\documentclass[letterpaper, 10 pt, conference]{ieeeconf}

\IEEEoverridecommandlockouts                              

\usepackage{cite}
\usepackage{amsmath,amssymb,amsfonts}
\usepackage{algorithmic}
\usepackage{graphicx}
\usepackage{textcomp}
\usepackage{xcolor}
\usepackage{url}
\usepackage{microtype}

\usepackage{hyperref}
\hypersetup{hidelinks}

\title{\LARGE \bf
A Multi-View 3D Telepresence System for XR Robot Teleoperation
} 

\author{
Enes Ulas Dincer$^{1}$, Manuel Zaremski$^{2}$, Alexandra Nick$^{2}$, Elias Wucher$^{1}$, Barbara Deml$^{2}$, Gerhard Neumann$^{1}$%
\thanks{$^{1}$ First, fourth, and sixth authors are with the Institute for Anthropomatics and Robotics, Karlsruhe Institute of Technology, 76137 Karlsruhe, Germany (e-mail: gerhard.neumann@kit.edu).}%
\thanks{$^{2}$ Second, third, and fifth authors are with the Institute of Human and Industrial Engineering, Karlsruhe Institute of Technology, 76137 Karlsruhe, Germany (e-mail: barbara.deml@kit.edu).}
}

\begin{document}

\maketitle
\thispagestyle{empty}
\pagestyle{empty}



\begin{abstract}

Robot teleoperation is critical for applications such as remote maintenance, fleet robotics, search and rescue, and data collection for robot learning. Effective teleoperation requires intuitive 3D visualization with reliable depth cues, which conventional screen-based interfaces often fail to provide.
We introduce a multi-view VR telepresence system that (1) fuses geometry from three cameras to produce GPU-accelerated point-cloud rendering on standalone VR hardware, and (2) integrates a wrist-mounted RGB stream to provide high-resolution local detail where point-cloud accuracy is limited. Our pipeline supports real-time rendering of approximately 75k points on the Meta Quest 3.
A within-subject study was conducted with 31 participants to compare our system to other visualisation modalities, such as RGB streams, a projection of stereo-vision directly in the VR device and point clouds without providing additional RGB information. Across three different teleoperated manipulation tasks, we measured task success, completion time, perceived workload, and usability. Our system achieved the best overall performance, while the Point Cloud modality without RGB also outperforming the RGB streams and OpenTeleVision.
These results show that combining global 3D structure with localized high-resolution detail substantially improves telepresence for manipulation and provides a strong foundation for next-generation robot teleoperation systems.

\end{abstract}



\section{Introduction}

Teleoperation remains crucial for remote robotic manipulation, especially for tasks that require dexterity and appropriate situational awareness~\cite{HumanRobotData2025,walker2024cyberphysical,Darvish2023TeleopHumanoids}. Conventional camera-based or monitor-based interfaces limit depth perception and spatial awareness~\cite{BaxterHomunculus2025,whitney2018rosreality,li2024realityfusion}. Many teleoperation systems that use virtual reality rely on multiple RGB camera streams, which provide high-resolution visual detail but lack the depth cues necessary for accurate spatial reasoning~\cite{cheng2024opentelevision, Luo2021TeleopUGV}. Conversely, point-cloud views have the potential to support better spatial understanding~\cite{jiang2025iris,zhao2025xrobotoolkit,Ni2017PointCloud}. Yet, the potential advantages of point-cloud visualzation interfaces have never been investigated by systematic user studies. Moreover, they often lack the fine visual detail required for grasping and contact-rich manipulation.  As a result, operators frequently face a trade-off between global scene understanding and high-resolution local detail~\cite{Mazeas.2025}. It also remains unclear how different visualization strategies affect performance in teleoperation with virtual reality.

We propose a new tele-presence system that extends coarse and rather noisy point-cloud visualizations \cite{jiang2025iris}  with precise, geometrically correctly arranged RGB information for the human operator. In our setup, point-clouds are visualized in the workspace of the tele-operation device (given by a leader robot that is moved via kinesthetic teachin) while the RGB stream of the wrist-camera is shown at the correct geometric position in the scene, allowing coarse geometric understanding of the scene using the point-cloud while the operator can perceive fine details of the objects to be manipulated via the RGB stream. We empower current VR point-cloud visualization systems with GPU accelerated rendering, enabling real-time visualization of up to 75{,}000 points. Using YOLOv11 for semantic filtering and GPU-accelerated point-cloud rendering, the system achieves real-time visualization of dense point clouds on the Meta Quest~3, enabling both global 3D understanding and local fine-detail perception.

Although various visualization paradigms have been explored, their relative advantages and disadvantages have not been systematically compared. In particular, no prior work has jointly evaluated the dominant visualization approaches used in modern VR teleoperation. We compare direct RGB multi-view streams (RGBs) of all cameras ~\cite{whitney2018rosreality,iyer2024open}, fused point-cloud (PC) ~\cite{hoang2021arviz,zea2020iviz,jiang2025iris,li2024realityfusion}, streaming of stereo camera feeds for depth perception as in OpenTeleVision~\cite{cheng2024opentelevision}, and our Point Cloud with wrist-mounted RGB (PC+RGB) visualization within a unified experimental setup. Furthermore, previous studies rarely consider multiple task types that challenge different perceptual and cognitive demands, such as insertion, alignment, and precision tracking. Consequently, the relationship between the visualization modality and operator performance, workload, and usability in VR robot teleoperation remains underinvestigated.

We focus on the central research question: \textit{How do different visualization modalities influence performance, workload, and user experience in teleoperation with virtual reality?} We evaluate this question in a controlled within-subject study where participants completed three representative manipulation tasks and experienced all four visualization modalities.

Our contributions are:
\begin{itemize}
    \item A GPU-accelerated telepresence system that combines multi-view point-cloud fusion with a high-resolution wrist-mounted RGB stream.
    \item A unified experimental framework that systematically compares four visualization modalities (RGBs, PC, PC+RGB, OT) across three manipulation tasks requiring insertion, alignment, and precision.
    \item Empirical results regarding performance, perceived workload, and subjective self-reports on the usability of different visualizations in VR teleoperation.
\end{itemize}

\section{Related Work}
\label{sec:related_work}

\subsection{XR Teleoperation Frameworks}

Teleoperation has emerged as a vital modality for robotic data collection, especially in manipulation tasks where remote dexterity and visual awareness are essential. Recent surveys highlight kinesthetic teaching and teleoperation as the dominant strategies for robot Learning from Demonstration (LfD)~\cite{barekatain2024roadmap}. Early platforms such as ROS Reality~\cite{whitney2018rosreality} and iviz~\cite{zea2020iviz} established foundational bridges between ROS-based robots and consumer VR/AR devices by streaming RGB and point cloud data into Unity or WebXR environments. Systems like ARviz~\cite{hoang2021arviz} extended these capabilities to mobile AR, offering generalized visualization tools rather than task-optimized control interfaces. Transitional frameworks such as AdaptiX~\cite{pascher2023adaptix} enhanced simulation fidelity and input device compatibility, but often fixed the visual modality to RGB or simulated mesh views, without isolating visualization effects on manipulation performance.
Recent works have shifted toward high-fidelity XR platforms designed for scalable data collection. Systems such as OPEN~TEACH~\cite{iyer2024open} and LeVR~\cite{weng2025levr} use the Meta Quest~3 to teleoperate single-arm and bimanual robots with stereo or passthrough RGB views, while IRIS~\cite{jiang2025iris} unifies simulation assets and real point clouds into shared XR scenes across Meta Quest~3 and HoloLens. Toolkits such as OpenVR~\cite{george2025openvr} and XRoboToolkit~\cite{zhao2025xrobotoolkit} streamline the setup and enable low-latency stereo streaming with ROS and OpenXR compatibility. Despite this progress, many frameworks commit to a single visualization mode and evaluate system-level performance, not the perceptual effects of alternate visual feedback strategies. Overall, current frameworks demonstrate that XR teleoperation is feasible, scalable, and suitable for data collection, yet, they rarely compare visualization modalities side-by-side under identical teleoperation conditions—a gap addressed by our empirical evaluation.

\subsection{Visualization Modalities for XR Telepresence}
\textit{(Stereo) RGB Streams.}
One of the most common modalities are RGB streams. RGB data can transmit high-resolution images, which are required for precise manipulation. However, these systems typically provide limited capabilities for the human operator to perceive the geometry of the scene. Systems such as OPEN TEACH and ROS Reality primarily rely on RGB video or passthrough views, which provide high resolution local detail but only implicit depth information~\cite{whitney2018rosreality,iyer2024open}. OpenTeleVision instead uses an actively controlled stereo RGB head camera on humanoid robots to deliver immersive ego-centric views and shows that stereoscopic, head-coupled vision can improve success rate and completion time on precise manipulation tasks compared to non-stereo alternatives~\cite{cheng2024opentelevision}. Yet, these systems often rely on a fixed camera position and therefore suffer from occlusions in case of cluttered scenes. 

\textit{Point Clouds.}
A second line of research treats XR primarily as a three dimensional visualization layer for robot-centric data, including depth and point clouds. ARviz and iviz use AR or VR headsets and mobile devices as general three dimensional viewers for ROS topics, rendering meshes, trajectories, and sensor streams in situ for debugging and human-robot interaction~\cite{hoang2021arviz,zea2020iviz}. IRIS extends this idea by streaming unified scene specifications, including simulated objects, robot models and real-time point clouds, into XR headsets to create digital twins that mix simulated and real content~\cite{jiang2025iris}.  Barone et al.\ propose a bandwidth-adaptive point cloud teleoperation pipeline with contention-aware scheduling and scalable transmission to maintain interactive frame rates under constrained network conditions~\cite{barone2025rtpc}. In parallel, volumetric and neural scene representations have been explored for telepresence. Reality Fusion fuses multi-modal RGB-D data into a unified volumetric scene for immersive mobile robot teleoperation~\cite{li2024realityfusion}, while Radiance Fields for Robotic Teleoperation and RIFTCast reconstruct radiance field or multi-view three-dimensional scenes to provide photorealistic free viewpoint XR telepresence~\cite{wildersmith2024radiancefields,zingsheim2025riftcast}. A recent virtual coexistence space study further compares VR, AR, and two-dimensional interfaces for remote industrial inspection, highlighting how fully three-dimensional XR views can improve situational awareness and collaboration~\cite{Mazeas.2025}.

\subsection{Human Spatial Perception for XR Teleoperation}
Human spatial perception is the fundamental prerequisite for the effective operation of any high-performance XR-based robot teleoperation system. It governs how accurately an operator can estimate distances, avoid collisions, maintain situational awareness, and trust the robot's actions~\cite{Moniruzzaman.2022}. When the remote scene is rendered with higher fidelity, operators are able to execute more precise control commands, thereby enhancing teleoperation performance across a broad range of complex tasks~\cite{Wang.2025}. According to the taxonomy proposed by Walker et al.~\cite{Walker.2023} for the evaluation of user interface design in XR, the selection of a visualization method is of critical importance. A single 2D camera is limited by a narrow field of view and low video quality, resulting in the operator's absence of stereoscopic depth cues~\cite{Girbés-Juan.2021}. Such ambiguity in object positions and poor end‑effector‑to‑target distance estimation necessitate a high degree of reliance on the operator's expertise. Multi-camera rigs have been shown to recover relative object poses and offer several viewpoints, thus partially compensating for a monocular sensor~\cite{Su.2023}. However, as telerobotic systems become more complex, reliance on multiple 2D streams for collision avoidance still limits control efficiency and increases the cognitive workload of the operator. However, binocular imaging or real-time point-cloud generation facilitates true depth perception and has been demonstrated to enhance manipulation performance~\cite{Su.2023}. Immersive XR interfaces show greater efficacy than traditional 2D displays in terms of spatial awareness, depth perception, and task accuracy, resulting in faster completion times and lower cognitive load~\cite{Su.2022}. The combination of depth-sensing hardware, point-cloud processing pipelines, and AR/VR overlays enhances spatial perception and the user experience, especially in scenarios requiring demanding manipulation~\cite{Mazeas.2025}. However, the question of whether stereoscopic vision and point-cloud integration within an XR environment can further enhance stereoscopic perception and task performance in tele-robotic operations remains an open research question~\cite{Su.2023}.     

\section{Stimuli and apparatures}
\label{sec:methods}

\subsection{Teleoperation Setup}
\label{subsec:teleop_setup}

\begin{figure}[h]
    \centering
    \includegraphics[width=\linewidth]{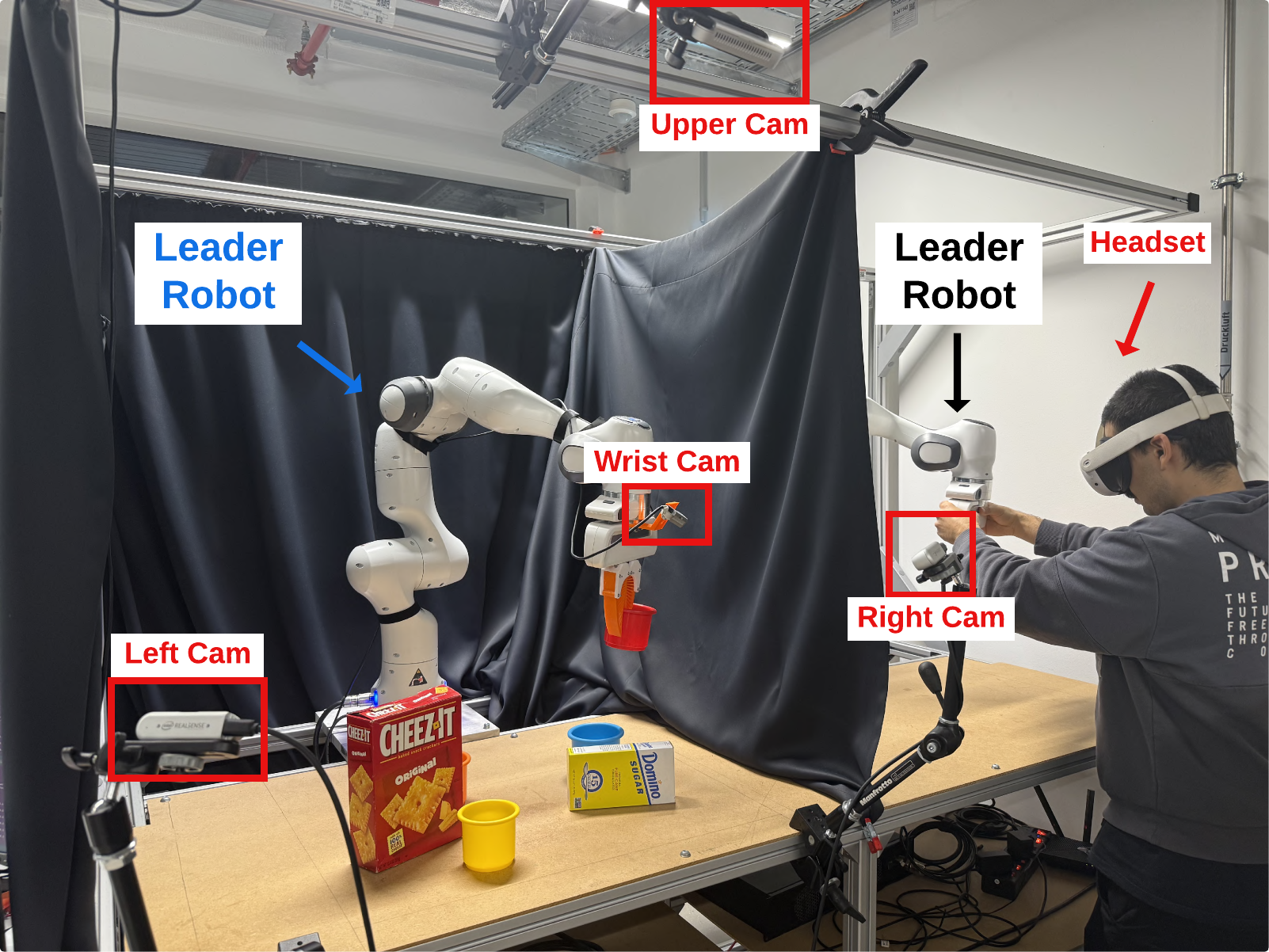}
    \caption{Leader–follower teleoperation system. One Panda arm acts as the leader, physically guided in gravity-compensation mode, while the second acts as the follower in the manipulation workspace. The workspace is observed by three static RGB-D cameras (left, right, upper) and a wrist-mounted RGB-D camera. The participant wears a Meta Quest~3 headset and experiences one of the four visualization conditions (RGBs, PC, PC+RGB, OT).}
   
    \label{fig:teleop_setup}
\end{figure}

Figure~\ref{fig:teleop_setup} shows the leader–follower teleoperation setup with two identical Franka Emika Panda arms. Participants physically guide the \emph{leader} arm in torque-based gravity-compensation mode, while the \emph{follower} arm executes the mirrored motion in the task workspace. Thus, teleoperation is performed directly via the physical robot: the operator stands next to the leader, grasps its end-effector, and performs the manipulation motion.

\textit{Sensing and XR Infrastructure.} The scene was observed by three static RGB-D cameras (Intel RealSense D415) mounted around the workspace: one on the left, one on the right, and one above the table. The three cameras were rigidly mounted and extrinsically calibrated with respect to the Panda base frame, providing overlapping coverage of the entire manipulation workspace from complementary viewpoints. A fourth RGB-D camera (Intel RealSense D435) was rigidly attached to the robot’s wrist and served as a high-resolution local RGB (and depth) source around the gripper.
A desktop PC collected and synchronized the streams from all four cameras and executed the perception pipeline that produces a fused, filtered point cloud (Section~\ref{subsec:vis_conditions}). The PC and the Meta Quest~3 head-mounted display were connected to the same wireless network; the PC sent compact per-point packets containing only the 3D coordinates and RGB values $(x,y,z,r,g,b)$ over WiFi. All subsequent GPU-based point-cloud rendering and compositing were performed directly on the Meta Quest~3 by a Unity application running on the headset. Once the Unity application received the point data, it instantiated the points using GPU shaders, enabling real-time visualization at the point budgets described below.

\subsection{Point-Cloud Generation and Fusion}
\label{subsec:pc_generation}
For visualizing the point-cloud, we run a synchronized pipeline for each frame. RGB and depth images from all three static RGB-D cameras are captured simultaneously, and each RGB image is processed by a YOLOv11 segmentation model trained on a custom dataset to detect and segment the robot, gripper, wrist-camera mount, and table, producing pixel-wise masks for these classes. We then generate 3D points only for \emph{unmasked} pixels by back-projecting the depth image using the camera intrinsics and assigning the corresponding RGB color to each point. This semantic filtering serves two purposes: (i) the robot and gripper are already rendered as accurate meshes from the Panda URDF in Unity, so including them in the point cloud would be redundant and consume part of the limited point budget; and (ii) removing the table and robot structure reduces visual clutter and allows the remaining points to focus on task-relevant objects, edges, and contact regions.

Each per-camera point cloud is transformed into the robot base frame using previously estimated extrinsic calibration matrices and merged into a single global cloud, which is then cropped with an axis-aligned 3D bounding box that tightly contains the tabletop and task objects, discarding all points outside the manipulation region. To satisfy the rendering constraints of the Meta Quest~3, we apply a voxel-grid downsampling followed by a statistical outlier removal filter: the voxel grid controls spatial point density, while the outlier filter suppresses isolated points and residual noise. The perception pipeline runs on a desktop PC equipped with an NVIDIA RTX~3060 GPU, and the resulting filtered point cloud contains approximately 75k colored points at an update rate of 10~Hz.

Finally, we serialize the point attributes into a compact $(x,y,z,r,g,b)$ representation and send this packet via WiFi to the Meta Quest~3. A Unity application on the headset receives the packet and reconstructs the point cloud on the GPU using shaders and instanced rendering. Without this on-device GPU-based rendering, real-time visualization would be limited to roughly 10k points; our pipeline therefore enables substantially denser point clouds under the same frame-rate constraints in comparison to \cite{jiang2025iris}.

\subsection{Visualization Modalities}
\label{subsec:vis_conditions}

\begin{figure*}[t]
    \centering
    \includegraphics[width=\textwidth]{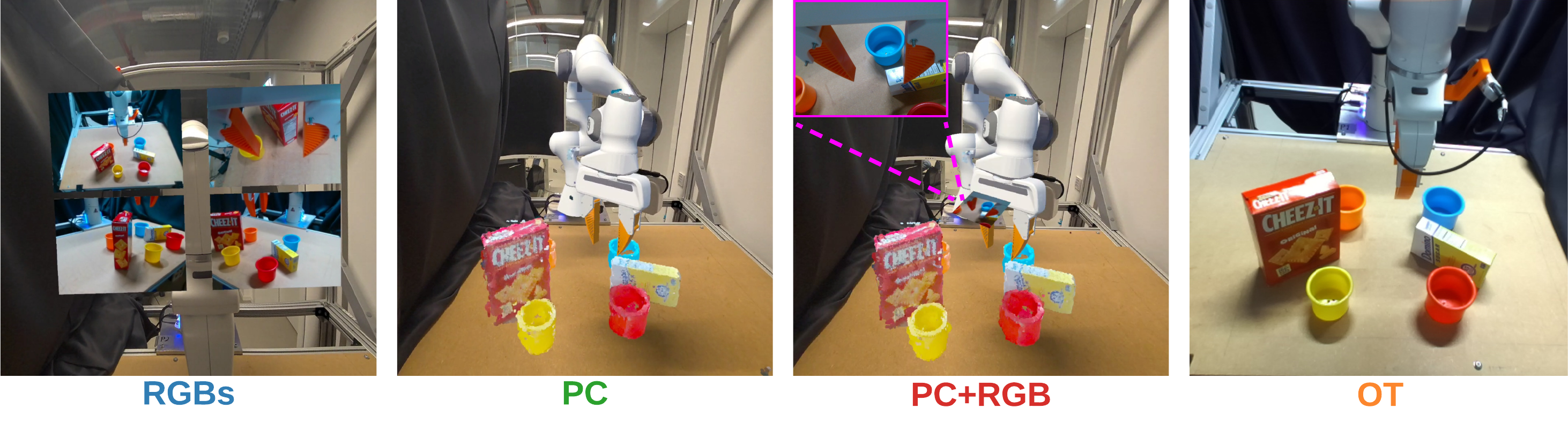}
    \caption{Visualization modalities evaluated in our VR teleoperation study, illustrated here for the cup–insertion task. From left to right:
(a) RGBs: four virtual screens showing left, right, top, and wrist RGB streams;
(b) PC: semantically filtered fused point cloud rendered together with the leader arm mesh;
(c) PC+RGB: the same fused point cloud augmented with a wrist–mounted RGB view rigidly attached to the end–effector. The magenta inset shows a zoomed crop of this wrist RGB stream, highlighting the available high–resolution local detail;
(d) OT: ego-centric stereo view from the OpenTeleVision setup.}
    \label{fig:visualizations_overview}
\end{figure*}

Our system supports four distinct visualization modalities that differ in how they present the scene to the operator. An overview is shown in Figure~\ref{fig:visualizations_overview}.

\textit{Four RGB Streams (RGBs).}
Users view four high-resolution RGB feeds (left, right, top, wrist) arranged as virtual floating screens in VR. While this setup offers sharp textures and edge cues from multiple angles, it provides only monocular depth cues and no explicit 3D scene structure.

\textit{Fused Point Cloud (PC).}
The PC condition renders only the semantically filtered point cloud from the three RGB-D cameras, aligned with the physical workspace in VR. The robot and gripper are shown as URDF meshes; the table is omitted except beneath task objects. This view provides a coherent 3D structure for assessing spatial relationships from arbitrary angles, though noise in depth perception, limited point budgets and downsampling reduce fine texture detail and edge clarity.

\textit{Point Cloud with Wrist-Mounted RGB (PC+RGB).}
PC+RGB combines the global spatial context of the fused point cloud with the fine detail of a close-up RGB stream. A floating window near the virtual end-effector displays the wrist-mounted RGB view. This hybrid setup supports both global 3D understanding and high-resolution local perception, in particularly important for grasp alignment and contact-rich manipulation.

\textit{OpenTeleVision (OT).}
Based on OpenTeleVision~\cite{cheng2024opentelevision}, this setup  shows a stereo RGB view from a headset-mounted ZED Mini camera fixed above the follower robot. Users see immersive first-person video from the follower’s perspective but lack direct visual feedback of their own arm or explicit 3D structure.


\section{Study Design} 
\label{sec:Experimental_Design}

\subsection{Sample}
The study included 31 healthy participants, 8 female, 22 male, and 1 non-binary/other between the ages of 20 and 33 ($M = 24.39,SD = 3.62$). The sample exhibited a STEM-oriented background, predominantly in the fields of computer science and engineering, and reported at least a vocational qualification or a Bachelor's degree. Nearly half of the participants (48.4\%) reported having no prior experience with gaming, VR, or simulator games, while 45.2\% reported using VR and simulators less than once per month. The ATI scale showed a moderately high affinity for technology (Cronbach’s $\alpha = 0.86$, $M = 4.44$, $SD = 0.78$, range 2.78–5.78 on a 1–6 scale).

\subsection{Procedure}
Participants were first informed about the study, data protection, and signed a consent form. They then completed questionnaires on demographics, technology affinity, and immersive tendency. A training session followed, during which the supervisor demonstrated gripper use and robot motion limits using a simple gripping task. Participants practiced the task both with and without the VR headset to familiarize themselves with the system.

\begin{figure}[tb]
\centerline{\includegraphics[width=0.49\textwidth]{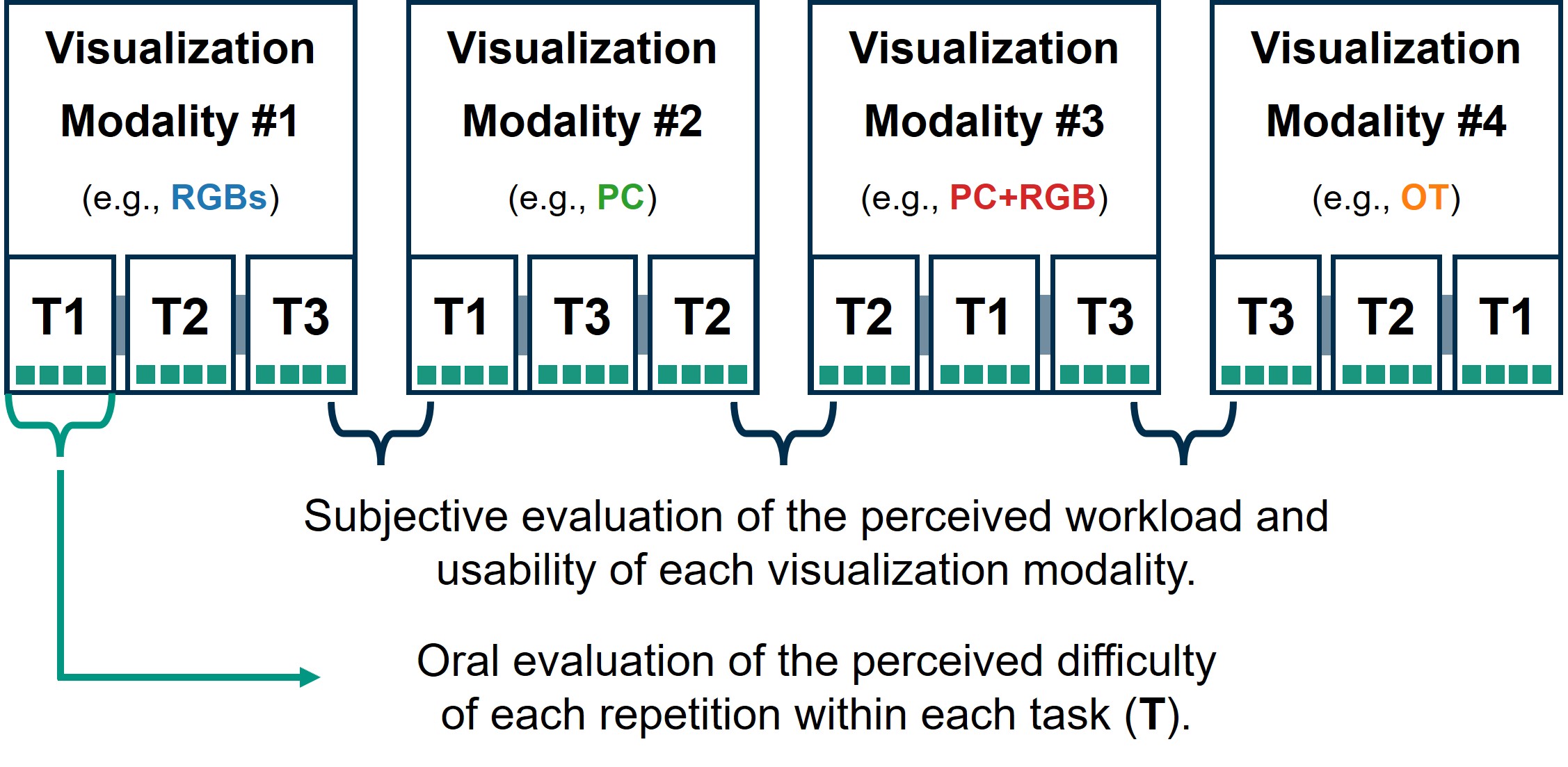}}
\caption{An overview of the study procedure, highlighting the elements of data collection and randomization of both the displayed visualization modalities and the manipulation tasks.}
\label{fig:procedure}
\end{figure}

Each participant completed all three manipulation tasks with four repetitions under each of the four visualization modalities (Figure~\ref{fig:procedure}), with randomized task and modality order. After each task, participants rated perceived difficulty. After each modality, they completed the NASA-TLX workload and VR usability questionnaires. At the end, participants ranked the four visualizations by preference and provided general feedback. The full session lasted approximately 95 minutes.

\subsection{Tasks}
\label{subsec:tasks}

\begin{figure}[t]
    \centering
    \includegraphics[width=1.00\linewidth]{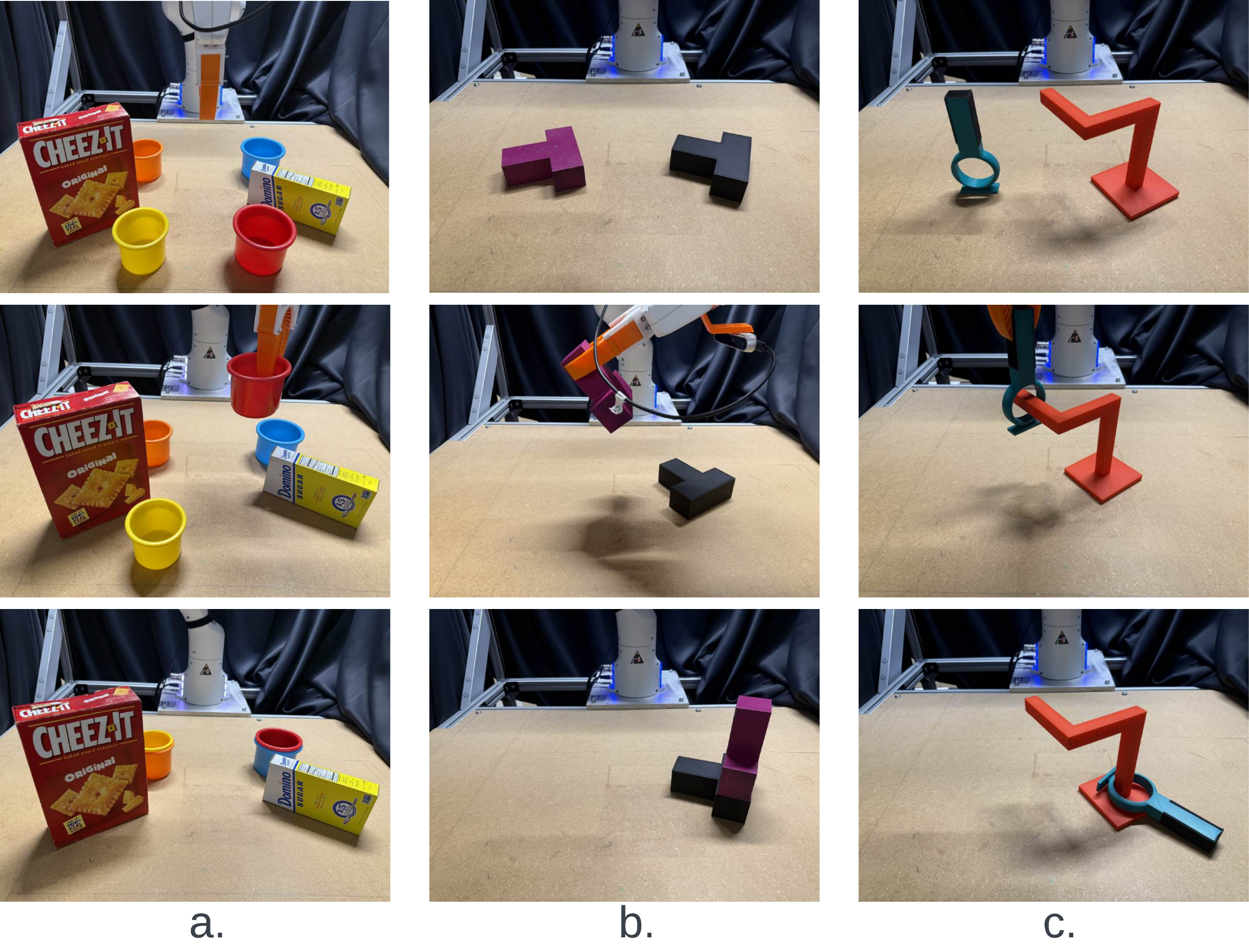}
    \caption{The three manipulation tasks used in the study, shown as start (top), intermediate (middle), and goal (bottom) states: (a) two sequential cup insertions with occluding obstacles, (b) T-shape stacking/assembly, and (c) wire-loop placement around an L-shaped stand.}
    \label{fig:tasks}
\end{figure}

We evaluate the four visualization conditions on three teleoperated manipulation tasks that cover insertion, alignment, and precision tracking. An overview of the tasks is shown in Figure~\ref{fig:tasks}.

\textit{Cup Insertion.}
This task involves two sequential insertions while avoiding nearby obstacles (Figure~\ref{fig:tasks}a). The scene contains four plastic cups (red, yellow, blue, orange) and two box-shaped obstacles (cereal and sugar boxes). At the start of each trial, the red and yellow cups start in a designated region; the blue and orange cups are partially occluded near the boxes. The participant needs to insert the red cup into the blue cup and the yellow cup into the orange cup, without dropping any cups or knocking over the boxes.

Participants completed two sequential cup insertions in a cluttered scene (Figure~\ref{fig:tasks}a): insert the red cup into the blue cup and the yellow cup into the orange cup. Two box obstacles (cereal and sugar) partially occlude the targets; trials fail if cups/boxes are knocked over or objects are dropped.

\textit{T-Shape Assembly.}
Participants must stack two T-shaped blocks (black and magenta) into a stable upright configuration (Figure~\ref{fig:tasks}b). The goal is to grasp the black T and place it on top of the magenta T such that the stems and crossbars are aligned to form a stable stacked configuration, requiring precise translational and rotational placement.

\textit{Wire-Loop.}
The wire-loop task is a precision placement task inspired by the classical wire-loop game (Figure~\ref{fig:tasks}c), involves a loop tool and an orange L-shaped stand. The tele-operated robot must grasp the loop, guide it over the stand, and place it around the base so that the handle lies flat on the table next to the stand. While contact with the stand is permitted, dropping either object results in failure, requiring precise control and deliberate placement.


\subsection{Measurements}
\label{subsec:tasks}
The effectiveness of teleoperation performance is quantified using objective and subjective metrics. These include the \textit{Task Success Rate}, which is a binary success indicator per trial of each manipulation task, \textit{Task Completion Time}, measured from the initial robot motion to task completion, \textit{perceived task difficulty}, using a single item and a 9-point Likert scale~\cite{revesz2016}, and the \textit{perceived workload} using the NASA-TLX~\cite{hart1988nasatlx}. In total, Participants have a time limit of 120 seconds to complete each task. Any attempts that exceed this time limit are recorded as unsuccessful. Furthermore, a set of questionnaires was conducted to gain more insight into the participants' experiences and to gather a self-report on the usability of the teleoperation setup. These include the affinity for technology interaction, as measured by the ATI-Scale~\cite{franke2019ATI}, the immersive tendency, as measured by the subscale of telepresence~\cite{witmer1998, scheuchenpflug2001}, and the usability of the teleoperation setup, measured by dimensions of effectiveness, efficiency, and satisfaction~\cite{Kim.2024}.  


\section{Results}
\label{sec:results}

All statistical analyses were carried out in R via RStudio~\cite{R2021,Rstudio2019}. For data handling and visualization, a range of established packages was employed, including tidyverse, dplyr, tidyr, ggplot2, ggsign, as well as psych, readr, ez, rstatix, and ggpubr.
Before hypothesis testing, the suitability of parametric procedures was assessed by testing normality with the Shapiro–Wilk test and checking sphericity with Mauchly’s test. Based on these results, Friedman tests were computed to determine whether differences between the visualization conditions were statistically significant. When significant overall effects were observed, pairwise post-hoc tests were executed using Wilcoxon signed-rank tests with continuity correction.

\subsection{Workload}
\begin{figure}[h]
    \centering
    \includegraphics[width=1.0\linewidth]{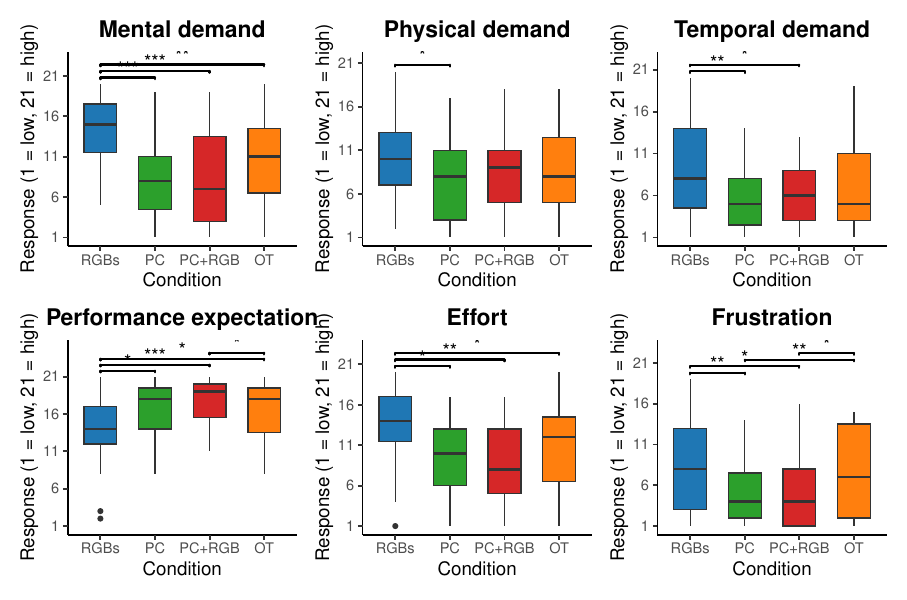}
    \caption{Boxplots of NASA--TLX ratings across the four visualization conditions (RGBs, PC, PC+RGB, OT). Higher values indicate higher subjective workload. Each subplot represents one workload dimension. Asterisks indicate the level of statistical significance (* p $<$ .05, ** p $<$ .01, *** p $<$ .001).}
    \label{fig:workload}
\end{figure}

Figure~\ref{fig:workload} summarizes the NASA--TLX ratings for all six workload dimensions across the four visualization conditions (RGBs, PC, PC+RGB, OT). Overall, the distributions show condition-dependent variation in mental, physical, and temporal demands, as well as in performance expectations, effort, and frustration. Descriptively RGBs generally elicited higher mental and temporal demand, whereas PC and PC+RGB tended to produce lower effort and frustration ratings. 
After checking the test assumptions and considering the sample size (n = 31), we decided to conduct non-parametric Friedman tests with Holm correction across the six workload dimensions.
The analysis revealed significant differences between the visualization conditions for all NASA–TLX dimensions. The significant results from the post-hoc tests are included in Figure~\ref{fig:workload}. 

\subsection{VR Usability} 

\begin{figure}[h]
    \centering
    \includegraphics[width=1.00\linewidth]{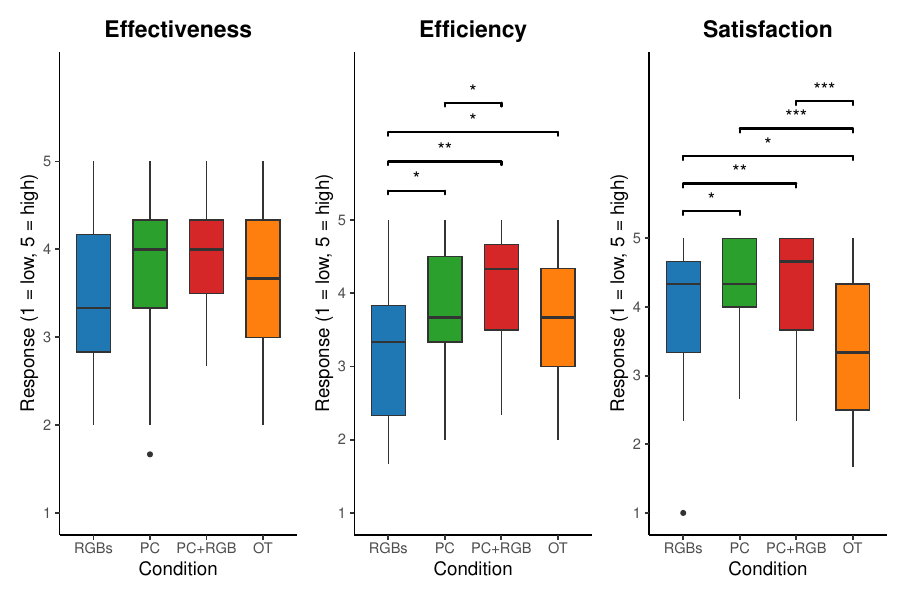}
    \caption{Boxplots of VR usability ratings (effectiveness, efficiency, satisfaction) across the four visualization conditions (RGBs, PC, PC+RGB, OT). Asterisks denote significant pairwise differences based on Holm-corrected Wilcoxon tests (* p  $<$ .05, ** p $<$ .01, *** p $<$ .001).}
    \label{fig:vrUsability}
\end{figure}

VR usability was assessed across three dimensions: effectiveness, efficiency, and satisfaction. Of the 12 Cronbach’s alpha values computed, 8 exceeded .70 and 4 were slightly lower—acceptable given the small number of items, and, hence, subscales were aggregated. Median scores across all dimensions were generally below the midpoint of the 1–5 scale, indicating overall positive usability. To examine whether usability differed across conditions, we conducted nonparametric Friedman tests because the score distributions were non-normal. The analyses revealed significant effects of the visualization condition for all three dimensions, indicating that effectiveness, efficiency, and satisfaction varied systematically across RGBs, PC, PC+RGB, and OT 
Post-hoc comparisons revealed no significant differences in perceived effectiveness. However, both efficiency and satisfaction showed significant pairwise differences, with PC and PC+RGB generally rated higher than RGBs and OT. Significant pairwise results are shown in Figure~\ref{fig:vrUsability}.


\subsection{Individual preference ranking}
\begin{figure}[h]
    \centering
    \includegraphics[width=0.95\linewidth]{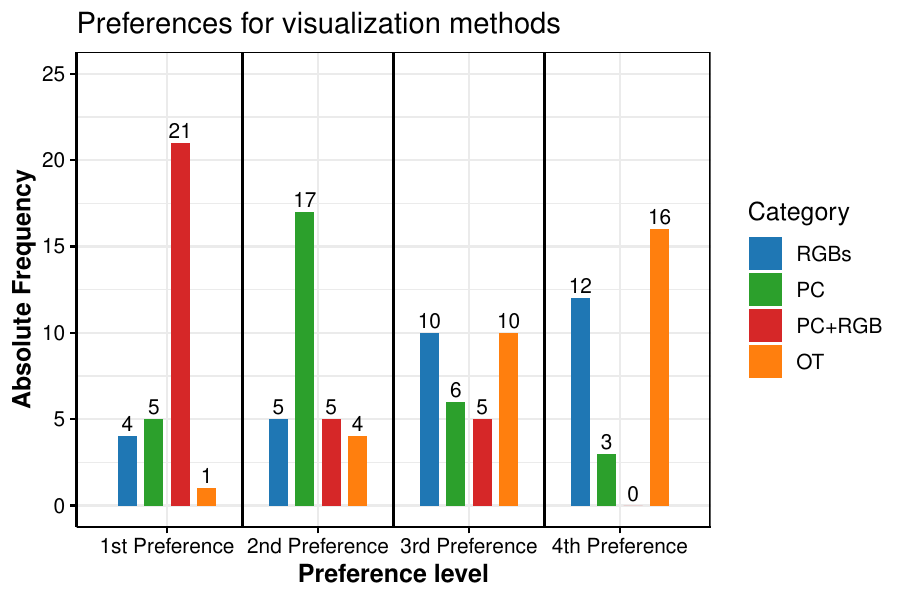}
    \caption{Participants’ subjective preference rankings for the four visualization methods across all preference levels.}
    \label{fig:personalPreference}
\end{figure}

At the end of the study, participants ranked the four visualization modalities in order of preference. PC+RGB was most frequently chosen as the first preference, followed by PC. OT was selected as the least preferred option by a small number of participants. Overall, these rankings indicate a clear preference for PC-based visualizations, with and without RGB augmentation (Figure~\ref{fig:personalPreference}).

\subsection{Performance analysis}

\begin{figure*}[h]
    \centering
    \includegraphics[width=1.0\linewidth]{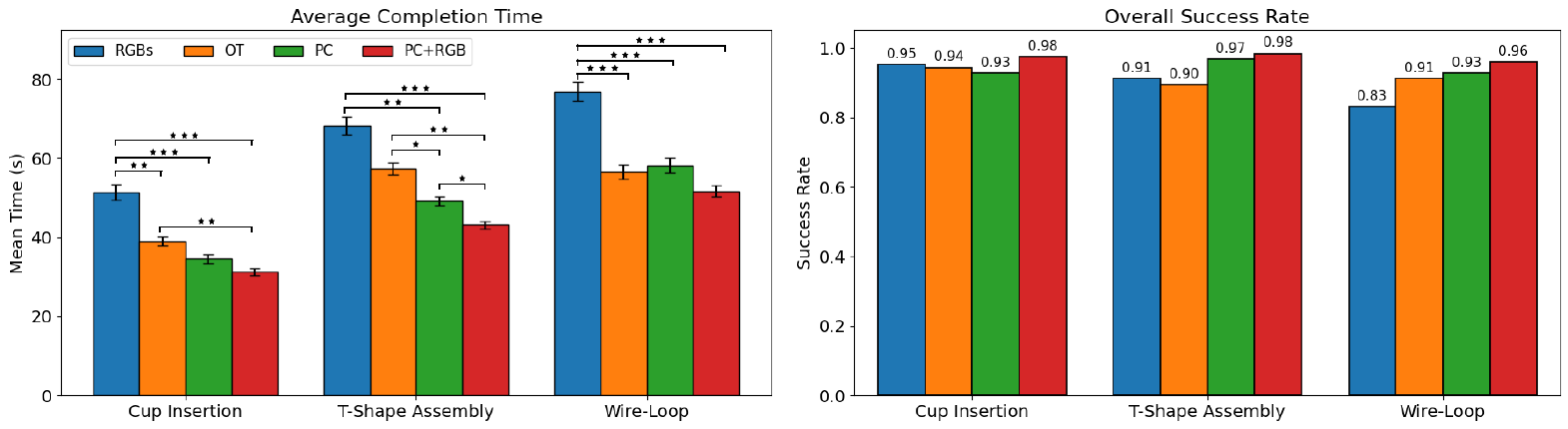}
    \caption{Task-wise performance across visualization conditions (RGBs, PC, PC+RGB, OT). Left: mean completion time for each task. Right: success rate for each task. Asterisks denote significant pairwise differences based on Holm-corrected Wilcoxon tests (* $p<.05$, ** $p<.01$, *** $p<.001$).}
    \label{fig:combinedPerformance}
\end{figure*}

\textit{Success rates} were high in all conditions but varied systematically with visualization (Figure~\ref{fig:combinedPerformance}). PC+RGB achieved the highest success rate, followed by PC and OT, whereas RGBs produced the most failures, indicating that PC-based views support more reliable execution than purely RGB-based ones.
For \textit{completion time}, a Friedman test indicated a significant effect of visualization condition. Holm-corrected pairwise Wilcoxon signed-rank tests (Figure~\ref{fig:combinedPerformance}) showed that RGBs was significantly slower than the depth-based conditions, while PC+RGB yielded the fastest completion times overall.

\section{Discussion}
\label{sec:discussion}

Our goal was to isolate how visualization alone affects VR-based teleoperation when the control interface is kept fixed. Across all measures, the point-cloud based conditions clearly outperformed purely RGB-based interfaces: PC+RGB achieved the best performance and usability, PC was consistently second, while RGBs and OT lagged behind.

A key explanation is the preservation of kinesthetic-teaching (KT)–style visuomotor co-location. In PC and PC+RGB, the fused scene of the follower arm and objects is rendered directly over the leader arm in VR, so participants experience their physical movements as acting on the virtual workspace. KT is widely recognized as effective for contact-rich manipulation and demonstration collection~\cite{barekatain2024roadmap}, and prior studies report superior performance compared to gamepads, motion controllers, and hand tracking~\cite{jiang2024comprehensive}. Our findings are consistent with this evidence: conditions that sustain KT-like co-location (PC, PC+RGB) produced lower workload and higher usability than RGBs and OT, where the mapping between hand motion and visual feedback is indirect.
Explicit three dimensional structure further distinguishes the conditions. RGBs provide sharp textures from four viewpoints, but users must infer depth from monocular cues and mentally integrate multiple camera views, which is cognitively demanding. The fused point cloud, in contrast, presents a single coherent three dimensional scene in which distances and clearances can be perceived directly; participants often described PC as helpful for “seeing where everything is’’ despite fuzzy edges. Adding the wrist-mounted RGB view in PC+RGB supplies the missing local detail, which explains why PC+RGB combines the lowest failure rates with the highest efficiency and satisfaction.
The OT condition shows that immersion alone is not sufficient. Although OT provides an ego-centric stereo view of the \emph{follower} robot that some participants found engaging, they never see the \emph{leader} arm they physically move, creating a proprioceptive–visual mismatch. This weaker sense of telepresence, together with reports of higher dizziness and workload, coincides with performance clearly below PC and PC+RGB, indicating that precise teleoperation benefits more from co-location and stable 3D structure, complemented by a foveated high-resolution view near the tool, than from ego-centric visual realism alone.

\section{Limitations}
\label{sec:Limitations}
The quality of our point cloud modalities is limited by the sensing and processing pipeline, which relies on three Intel RealSense D415 cameras, YOLOv11 based masking, and voxel grid downsampling under a strict point budget on the Meta Quest~3. Participants sometimes described the point clouds as visually fuzzy at edges, suggesting that better sensors or reconstruction methods could improve spatial precision without changing the qualitative advantage of PC based conditions.
Our sample consisted mainly of young adults with STEM backgrounds and little prior VR experience, which does not reflect all potential teleoperators such as industrial technicians or remote service staff. Future work should include more diverse user groups and longer, more demanding teleoperation sessions to test how robust the benefits of PC and PC+RGB are in real world deployments.

\section{Conclusion}
\label{sec:Conclusion}

We presented a multi-view 3D telepresence system for VR-based robot teleoperation that fuses three RGB-D cameras into a GPU-accelerated point cloud and augments this global structure with a high-resolution wrist-mounted RGB stream. On a standalone Meta Quest~3 headset, the system renders approximately 75k colored points in real time. In a within-subject study with 31 participants, we compared four visualization modalities (RGBs, PC, PC+RGB, OT) across three manipulation tasks and measured task success, completion time, workload, and usability. PC+RGB consistently achieved the best overall performance, with PC also outperforming RGBs and OT in most metrics.
These findings demonstrate that combining global 3D scene structure with localized high-resolution detail substantially improves telepresence for manipulation, compared to both traditional multi-camera RGB interfaces and immersive stereo video alone. For designers of XR teleoperation frameworks, our results argue for hybrid visualizations that explicitly represent depth while providing foveated RGB detail near the tool. Future work will extend this approach to higher-fidelity scene representations and more diverse task domains, and investigate adaptive visualization strategies that dynamically allocate visual bandwidth where operators need it most.

\section*{Ethics Statement}
The user study involved human participants and was conducted in accordance with institutional ethical guidelines. Ethical approval was obtained from the responsible ethics committee prior to the study. All participants provided informed consent before participation.


\bibliographystyle{IEEEtran}
\bibliography{references}

\end{document}